\ifx\XeTeXversion\undefined
  \ifdefined\pdfoutput
    \pdfoutput=1
  \fi
\fi
\documentclass[10pt]{article}

\usepackage[margin=1in]{geometry}
\usepackage{amsmath}
\usepackage{amssymb}
\usepackage{booktabs}
\usepackage{graphicx}
\usepackage{caption}
\usepackage{hyperref}
\usepackage{natbib}
\usepackage{enumitem}
\usepackage{xcolor}

\hypersetup{
  colorlinks=true,
  linkcolor=black,
  citecolor=blue,
  urlcolor=blue
}
\makeatletter
\@ifundefined{doi}
  {}
  {}
\makeatother
\setcitestyle{authoryear,round,aysep={,},yysep={;}}
\setlist[itemize]{leftmargin=*, itemsep=2pt, topsep=3pt}
\title{AutoTail-BSFGM: Class-Balance-Aware Fine-Tuning for\\
Chinese Scholarly Text Classification}

\author{
Anling Xiang\textsuperscript{1},
Yuwen Yang\textsuperscript{2},
Yang Shen\textsuperscript{3,4}\\
\textsuperscript{1}Department of Intelligent Communication, School of Journalism and Communication,\\
Minzu University of China, Beijing, China\\
\textsuperscript{2}ZeeLin (Beijing) Technology Co., Ltd., Beijing, China\\
\textsuperscript{3}School of Journalism and Communication, Tsinghua University, Beijing, China\\
\textsuperscript{4}College of AI, Tsinghua University, Beijing, China\\
\small ORCID: Anling Xiang \href{https://orcid.org/0000-0003-1690-1586}{0000-0003-1690-1586}\\
\small Yuwen Yang \href{https://orcid.org/0009-0002-3084-6720}{0009-0002-3084-6720};
Yang Shen \href{https://orcid.org/0000-0003-4814-9018}{0000-0003-4814-9018}
}

\date{\today}

\begin{document}
\maketitle

\begin{abstract}
Scholarly text classification supports literature organization, research intelligence, subject indexing, and large-scale science-of-science analysis. In Chinese scholarly corpora, however, disciplinary labels are often imbalanced, semantically adjacent, and sensitive to input granularity: abstracts contain rich topical evidence, whereas titles provide short and sometimes ambiguous signals. These properties make it insufficient to report only a single validation accuracy from ordinary fine-tuning. This paper studies a conservative method for robust and class-balance-aware Chinese scholarly text classification. The proposed method, AutoTail-BSFGM, combines three training-time components: an automatically gated tail-prior adjustment, a weak Balanced Softmax auxiliary loss, and Fast Gradient Method adversarial regularization. The method changes only the fine-tuning objective and training procedure. At inference time it uses the same single base-size encoder and linear classifier as the corresponding label-smoothed baseline.

We evaluate AutoTail-BSFGM on two CSL-based scholarly classification tasks: a constructed abstract-to-discipline task with 67 discipline labels and the official CSL title-to-category task with 13 broad categories. On the primary abstract-to-discipline task, AutoTail-BSFGM improves over label-smoothed Chinese RoBERTa-WWM by 1.42 validation accuracy points and 0.49 lockbox accuracy points across three seeds. With MacBERT-base, it improves validation accuracy by 0.83 points and lockbox accuracy by 0.49 points, with a pooled paired McNemar signal on validation ($p=0.023$). On the title-to-category task, the method improves validation accuracy by 0.70 points, validation macro-F1 by 1.33 points, and validation balanced accuracy by 2.64 points, while lockbox accuracy is neutral/slightly negative and lockbox balanced accuracy improves by 1.22 points.

The evidence supports a bounded contribution rather than a leaderboard claim: AutoTail-BSFGM improves class-balance-sensitive behavior and yields consistent gains on abstract-based scholarly classification, but it does not uniformly improve every accuracy measure on every split. We provide run identifiers, generated tables, prediction-derived significance checks, and an artifact manifest to make the result auditable and failure-aware.
\end{abstract}


\noindent\textbf{Code and data:} \url{https://github.com/thu-nmrc/autotail-bsfgm-scholarly-classification}

\section{Introduction}

The organization of scientific literature is a core problem in information science. Bibliographic databases, digital libraries, and research intelligence systems rely on reliable classification to support subject indexing, search, trend detection, portfolio analysis, and the study of science itself. Traditional bibliometric and science-of-science work has emphasized citations, publication networks, and field structures \citep{garfield1972citation,narin1976evaluative,bornmann2015growth,fortunato2018science}. More recent systems increasingly combine metadata, abstracts, full text, and language models to characterize research aims, topics, and disciplinary relations \citep{wu2025scaling,beltagy2019scibert,lo2020s2orc,singh2023scirepeval,zhang2024scientificllm}. In this setting, text classification is not only a benchmark exercise. It shapes which documents are retrieved, how emerging areas are grouped, and how small but important fields are represented in downstream analysis.

Chinese scholarly text introduces several practical difficulties. First, disciplinary taxonomies are long-tailed. Broad engineering and science categories may dominate, while smaller fields have far fewer examples. Second, boundaries between adjacent disciplines are often blurred at the language level. A paper about medical imaging, biomedical engineering, and computer vision can contain lexical evidence for multiple categories. Third, input granularity matters. Abstracts provide richer evidence but can include method, object, and application terms at once; titles are shorter and may be more brittle. These properties make ordinary accuracy-optimized fine-tuning potentially misleading: a method can improve the dominant classes while leaving rare categories under-served, or improve balanced metrics while slightly reducing overall accuracy on a protected split.

Pretrained language models provide strong baselines for scholarly classification. BERT \citep{devlin2019bert}, RoBERTa \citep{liu2019roberta}, Chinese whole-word-masking models \citep{cui2019wwm}, and MacBERT \citep{cui2020revisiting} have made single-encoder fine-tuning a practical default. Recent text-classification work has also explored in-context learning, label verbalization, prompt-based augmentation, and LLM-assisted classification \citep{edwards2024language,yu2024sciprompt,chen2023promptda}. These directions are important, but they can require additional prompt engineering, external model calls, or generated data. The present study instead asks whether a local, same-encoder fine-tuning objective can improve balance-aware behavior. Prior-aware long-tail methods, including Balanced Softmax \citep{ren2020balanced}, logit adjustment \citep{menon2021logit}, and label-distribution-aware margins \citep{cao2019ldam}, offer useful principles, but direct adoption can produce head-tail tradeoffs. Robust fine-tuning methods, including adversarial training \citep{goodfellow2015explaining,miyato2016adversarial,zhu2020freelb,jiang2020smart}, can improve local stability, but they also add training cost and may not address class imbalance by themselves.

This paper studies a narrow but practically important question: can a small, auditable modification to the fine-tuning objective improve class-balance-aware behavior in Chinese scholarly text classification without larger encoders, external private data, ensemble inference, or task-specific pretraining? We propose AutoTail-BSFGM, a training-time method that combines an imbalance-gated tail-prior adjustment, a weak Balanced Softmax auxiliary objective, and Fast Gradient Method (FGM) adversarial regularization. The method is intentionally conservative. It keeps the inference model unchanged and evaluates gains against same-encoder label-smoothed baselines under fixed seeds and a protected lockbox split.

The contribution is also methodological. The study is designed to avoid over-claiming from a single favorable validation split. We evaluate two tasks derived from CSL \citep{li-etal-2022-csl}: a 67-label abstract-to-discipline task and a 13-label title-to-category task. We report accuracy, macro-F1, and balanced accuracy, and we include prediction-derived paired McNemar checks. The resulting evidence is positive but bounded. The method consistently improves the primary abstract task under two encoders and improves balanced metrics in the title task, but it does not uniformly improve lockbox accuracy in the title setting. This mixed pattern is important: it suggests a real class-balance signal while preventing an exaggerated claim of universal superiority.

The paper makes four contributions:
\begin{itemize}
  \item It formulates Chinese scholarly classification as a class-balance-sensitive information science problem rather than only a generic NLP benchmark.
  \item It introduces AutoTail-BSFGM, a single-model fine-tuning method combining gated tail-prior correction, weak Balanced Softmax supervision, and FGM adversarial regularization.
  \item It evaluates the method on abstract-level and title-level CSL tasks using three seeds, protected lockbox checks, significance tests, and training-cost accounting.
  \item It provides category-level analysis showing that improvements are most naturally interpreted as balance-aware gains rather than unconditional leaderboard gains.
\end{itemize}

The framing of the paper is deliberately different from a model-release report. Many classification papers in NLP emphasize a new score on a benchmark table. Such a framing is useful when the task definition, evaluation metric, and comparison set are stable. Scholarly classification for information science has a broader set of concerns. A classifier can be useful if it improves the representation of smaller fields in an indexing workflow, even if the absolute leaderboard position remains below much larger systems. Conversely, a higher validation score can be misleading if it is produced by sacrificing protected categories or by overfitting to a convenient split. For this reason, the study emphasizes auditable improvement under a fixed protocol: same backbone, same data split, same seeds, explicit cost, and a protected lockbox.

This claim boundary is central to the paper. AutoTail-BSFGM is not presented as a new pretrained language model, a general long-tail theory, or a universal state-of-the-art method. It is a reproducible fine-tuning mechanism for a specific but important class of tasks: Chinese scholarly text classification with imbalanced disciplinary labels. The contribution is meaningful only if the reader can see both the positive signal and the negative boundary. The abstract task supplies the positive evidence; the title task supplies a useful stress test that prevents over-generalization.

The intended reader is therefore both methodological and applied. For machine-learning readers, the paper isolates a training objective and evaluates it under same-encoder controls. For information science readers, it asks whether the resulting classifier behaves more responsibly across disciplinary categories. This dual framing explains why the paper includes formal loss definitions, run-level reproducibility details, category-level interpretation, and explicit limitations rather than only a single benchmark score or a visually attractive but weakly audited claim. It also makes the evidence package easier for external readers to inspect, repeat, and challenge.

\section{Related Work}

\subsection{Scholarly Text Classification and Research Intelligence}

Information science has long studied how scientific publications can be organized, classified, and evaluated. Citation analysis and bibliometrics supplied early quantitative tools for journal evaluation and research assessment \citep{garfield1972citation,narin1976evaluative}. Later work developed field-level and publication-level classification systems that combine citation relations with textual evidence \citep{waltman2012new,boyack2014characterizing}. These approaches remain important because scientific categories are not merely labels for prediction; they are interfaces through which users navigate knowledge domains.

The growth of scientific publication volume has increased the need for automated classification and analysis \citep{bornmann2015growth,fortunato2018science}. Scholarly NLP has responded with corpora, language models, and pipelines for tasks such as citation intent, information extraction, discourse analysis, summarization, and knowledge graph construction \citep{cohan2018discourse,ammar2018construction,luan2018multi,lo2020s2orc}. Domain-specific pretrained encoders such as SciBERT show that scientific language has enough specialized structure to benefit from tailored modeling \citep{beltagy2019scibert}. More recent evaluation work, including SciRepEval, emphasizes that scientific document representations should be assessed across multiple downstream tasks rather than a single isolated score \citep{singh2023scirepeval}. Recent JASIST-style work further demonstrates the relevance of language models to large-scale classification of scientific aims and societal orientations \citep{wu2025scaling}.

Recent work has also made scientific-document classification itself a direct research object. FoRC4CL develops a fine-grained field-of-research taxonomy and annotated corpus for NLP articles \citep{ahmad2024forc4cl}. Scientific abstract classification has been studied through an LLM-assisted workflow \citep{sakhrani2024artificial}, while other studies examine zero-shot field classification and efficient few-shot multi-label classification for scientific documents \citep{gelles2024multi,schopf2024efficient}. Together, these studies reinforce the relevance of scholarly classification to indexing, navigation, and research-assisting systems.

Chinese scientific literature has different language and metadata conditions from English scientific corpora. CSL provides a large-scale Chinese scientific literature dataset and benchmark tasks that include titles, abstracts, keywords, and category metadata \citep{li-etal-2022-csl}. Its scale makes it suitable for modern pretrained models, while its labels expose common scholarly-classification challenges: imbalanced categories, heterogeneous disciplinary boundaries, and varying input lengths. Our work uses CSL not as a leaderboard endpoint but as a controlled environment for examining whether a class-balance-aware fine-tuning method can improve scholarly classification behavior.

Recent work has also moved beyond ordinary encoder fine-tuning. Fine-grained scientific topic classification can exploit domain terminology and label verbalizers, as shown by SciPrompt \citep{yu2024sciprompt}. Broader text-classification studies have compared fine-tuning with in-context learning and found that prompting alone is not always a substitute for task-specific training \citep{edwards2024language}. These findings motivate our same-backbone design: the goal is not to replace current LLM-oriented methods, but to isolate a reproducible training-objective intervention that can be evaluated without external model calls or generated labels.

This position connects the present work to JASIST-style research on methodological support for scientific knowledge organization. The relevant question is not only whether a model can classify text, but whether the model's behavior is interpretable enough for use in bibliometric and research-intelligence settings. Prior studies that combine text mining with bibliographic structure show that classification choices can affect field delineation and downstream measurement \citep{waltman2012new,boyack2014characterizing}. A small shift in category assignments may change how a topic is counted, which papers are retrieved for expert review, or which emerging field appears to be growing. Therefore, an evaluation that reports only average accuracy is incomplete for the intended application.

\subsection{Long-Tailed Learning and Class-Prior Correction}

Class imbalance is a long-standing problem in machine learning \citep{he2009learning}. In long-tailed settings, empirical risk minimization tends to favor frequent classes, which may increase overall accuracy while reducing performance on minority classes. Prior studies have explored resampling, reweighting, margin adjustment, focal loss, and post-hoc prior correction \citep{lin2017focal,cao2019ldam,buda2018systematic}. More recent methods incorporate class priors directly into logits or softmax normalization. Balanced Softmax adjusts the normalization term according to label frequencies \citep{ren2020balanced}, while logit adjustment provides a simple and theoretically motivated prior-aware correction \citep{menon2021logit}. For NLP specifically, \citet{henning2023survey} survey methods for addressing class imbalance in deep-learning-based natural language processing, organizing prior work around sampling, data augmentation, loss functions, staged learning, and model design. Newer text-classification work further explores label semantics, retrieval-style label matching, and LLM-driven data augmentation for long-tail labels \citep{chen2023promptda,wang2023pesco,zhang2023longtailxml,lu2026dealt}. These methods clarify that the training distribution, label meaning, and decision rule should not be treated as independent.

However, scholarly classification differs from some canonical long-tail benchmarks. Labels can be semantically close, not merely visually or syntactically distinct. Tail labels may share terminology with head labels, and a strong correction may overcompensate. In preliminary exploration, fully applying a prior correction can improve one split while degrading another. This motivates a more constrained design: AutoTail applies prior adjustment only when the observed label distribution is sufficiently imbalanced and only to a tail subset. Balanced Softmax is used as a weak auxiliary loss rather than a replacement for the main objective. The design is deliberately cautious because the intended use case is research intelligence, where unstable category shifts can mislead downstream interpretation.

A second issue is that long-tail gains can be reported in incompatible ways. Some studies emphasize mean per-class accuracy, others report macro-F1, and benchmark tables often use overall accuracy. These metrics answer different questions. Overall accuracy reflects expected correctness under the observed distribution; macro-F1 and balanced accuracy ask whether the classifier treats classes more evenly. In a scholarly database, both views are relevant. A production indexing system cannot ignore majority-class accuracy, but an analyst studying small fields cannot accept a classifier that performs well only on head categories. This is why the present paper reports all three metrics and treats disagreement among them as evidence rather than noise.

\subsection{Robust Fine-Tuning of Pretrained Language Models}

Pretrained language models have become standard for text classification \citep{devlin2019bert,liu2019roberta}. Chinese NLP further benefits from whole-word masking and MacBERT-style pretraining improvements that better fit Chinese tokenization and semantic masking conditions \citep{cui2019wwm,cui2020revisiting}. These encoders are strong enough that many apparent gains disappear when compared against an appropriate same-encoder baseline. For this reason, the present study compares AutoTail-BSFGM against label-smoothed baselines using the same encoder, training budget, and seeds.

Robust fine-tuning addresses another source of instability. Adversarial training perturbs embeddings or input representations in the direction that most increases the loss, then optimizes the model to resist that perturbation \citep{goodfellow2015explaining,miyato2016adversarial}. Variants such as FreeLB and SMART improve robustness and generalization for language understanding tasks \citep{zhu2020freelb,jiang2020smart}. R-Drop uses consistency regularization across dropout masks \citep{liang2021rdrop}. These methods are relevant to scholarly texts because small lexical changes can influence category assignment, especially in short titles or abstracts that combine method and application terms. AutoTail-BSFGM uses FGM as a lightweight regularizer that does not alter inference-time computation.

The choice of FGM is pragmatic. Stronger adversarial or consistency methods may be more powerful, but they can also introduce more hyperparameters, longer training cycles, or additional implementation complexity. The goal of this paper is not to exhaust the robust fine-tuning design space. It is to test whether a compact combination of prior-aware and robustness-aware mechanisms can produce a measurable improvement under a strict same-backbone comparison. This design choice makes the evidence easier to audit and the method easier to reproduce on ordinary hardware.

\section{Data and Methodology}

\subsection{Data Construction and Task Definitions}

\begin{figure}[t]
\centering
\includegraphics[width=\linewidth]{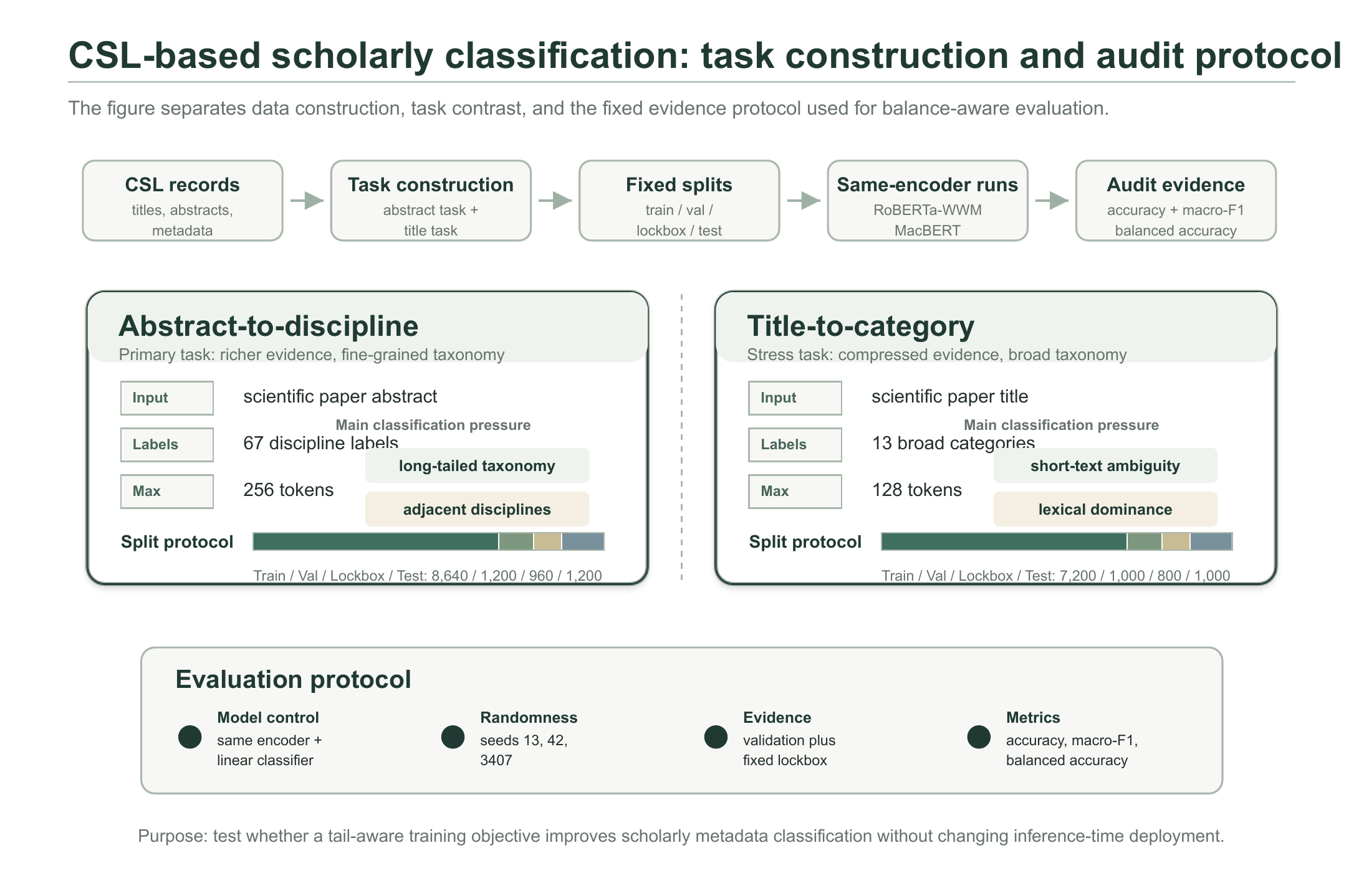}
\caption{Task overview for the two CSL-based scholarly classification settings. The primary abstract-to-discipline task uses longer abstracts and 67 discipline labels; the title-to-category task uses shorter titles and 13 broad categories. Validation and lockbox splits are reported separately to reduce the risk of selecting a method that only wins on a single development split.}
\label{fig:task_overview}
\end{figure}

Figure~\ref{fig:task_overview} summarizes the two evaluation tasks. The primary task is a CSL-derived abstract-to-discipline classification problem. We construct it from instruction-style CSL records by retaining abstract-based examples, removing title-only or keyword-only instructions, filtering examples with insufficient textual content, and keeping labels with sufficient support. The resulting task maps scientific abstracts to 67 discipline labels. The prepared local split contains 8,640 training examples, 1,200 validation examples, 960 lockbox examples, and 1,200 test examples. The maximum sequence length is 256.

The second task is CSL title-to-category classification. It maps scientific titles to 13 broad categories and uses a shorter maximum sequence length of 128. Compared with the abstract task, it has fewer labels but a more compressed input. It therefore tests a different failure mode: whether a method tuned for class-balance-sensitive abstract classification transfers to a title-level task where individual words may dominate the decision.

The abstract-to-discipline task is closer to a subject-indexing scenario. The model sees enough text to infer research objects, methods, and application domains, but the 67-label taxonomy creates many opportunities for near-miss errors. The title-to-category task is closer to a lightweight metadata-enrichment scenario. It is attractive because titles are widely available, but it is also harder to interpret: a title may describe a method without naming the target discipline, or it may use broad terms that appear across fields. Using both tasks allows the paper to distinguish between a method that only exploits abstract length and a method that has a more general class-balance effect.

Data construction follows two conservative principles. First, no private or manually relabeled data are introduced. All examples come from CSL-derived public records or local conversions of official CSL tasks. Second, the protected lockbox split is separated before final comparison. The lockbox is not a substitute for an official test server, but it reduces the risk of selecting a configuration only because it is favorable on validation. In future journal submission, the same protocol can be extended to additional datasets and official submissions; the present arXiv version records the initial evidence package.

\begin{table}[t]
\centering
\caption{Dataset statistics used in the experiments. The lockbox split is held out from model selection and used as a protected check.}
\label{tab:dataset_stats}
\small
\setlength{\tabcolsep}{3.5pt}
\begin{tabular}{@{}llrrrrrr@{}}
\toprule
Task & Input & Labels & Train & Val. & Lockbox & Test & Max len. \\
\midrule
CSL abstract-to-discipline & Scientific abstract & 67 & 8640 & 1200 & 960 & 1200 & 256 \\
CSL title-to-category & Scientific title & 13 & 7200 & 1000 & 800 & 1000 & 128 \\
\bottomrule
\end{tabular}

\end{table}

This two-task design is intentionally modest. It does not claim to cover all Chinese scholarly classification scenarios. Instead, it provides a primary abstract setting and a second title setting with different label and input properties. The goal is to determine whether the method has a stable balance-aware signal before making stronger claims.

\subsection{Evaluation Protocol}

The evaluation protocol separates validation and lockbox evidence. Validation is used for routine model comparison. Lockbox is reserved for protected checking after the candidate configuration is fixed. This distinction matters because small text-classification improvements can be artifacts of split-specific tuning. For each main comparison, we run three seeds: 13, 42, and 3407. We report mean accuracy, macro-F1, balanced accuracy, and mean training time. Accuracy remains the headline metric because it is commonly used in CSL classification benchmarks, but macro-F1 and balanced accuracy are essential for interpreting a method designed around class imbalance.

The seed design addresses randomness from initialization, data order, dropout, and low-level training nondeterminism. A single seed is useful for smoke testing but weak evidence for a method claim. Three seeds are not exhaustive, but they are sufficient to reveal whether a reported effect is directionally stable under ordinary fine-tuning variance. The results should therefore be read as an initial reproducibility check rather than a final statistical certificate. A stronger journal version should use more seeds or repeated split evaluations if computational budget allows.

We also compute paired McNemar tests from prediction files for the MacBERT comparisons. The test counts examples that are corrected only by the baseline and examples corrected only by the candidate. It therefore measures whether two classifiers make systematically different errors on the same examples. We treat the result as supporting evidence rather than a complete decision rule, following the broader caution that classifier comparison should consider multiple sources of variation \citep{dietterich1998approximate}. A method that improves validation accuracy but fails lockbox, balanced metrics, or cost checks should not be described as a general breakthrough.

\subsection{AutoTail-BSFGM Framework}

\begin{figure}[t]
\centering
\includegraphics[width=\linewidth]{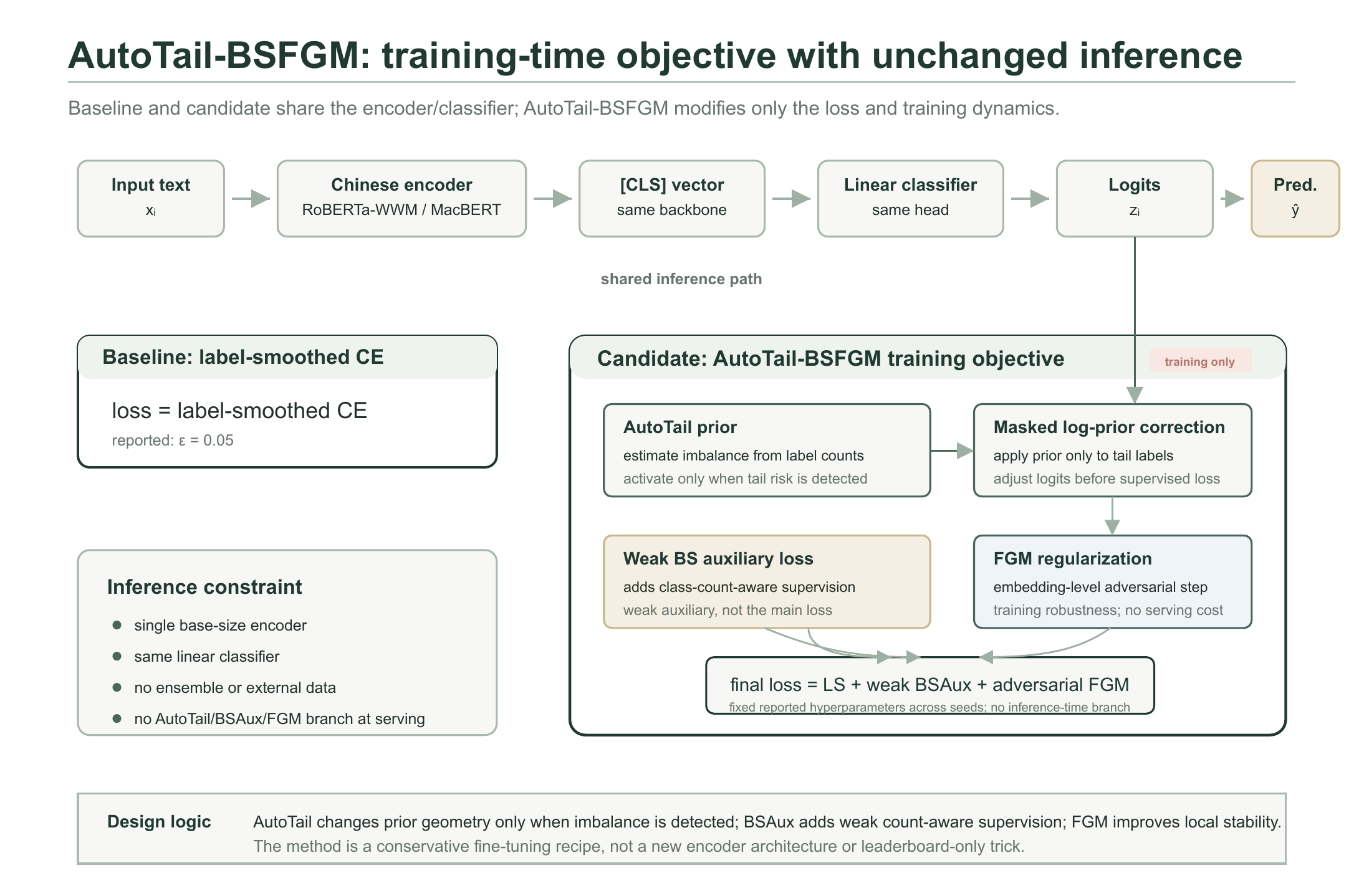}
\caption{AutoTail-BSFGM framework. The baseline uses label-smoothed cross-entropy. AutoTail-BSFGM adds an imbalance-gated tail prior to the logits, uses a weak Balanced Softmax auxiliary objective, and applies FGM adversarial regularization during training. Inference remains a single encoder and linear classifier.}
\label{fig:method_framework}
\end{figure}

The proposed method is shown in Figure~\ref{fig:method_framework}. Given an input text $x_i$ and label $y_i \in \{1,\ldots,K\}$, a pretrained Chinese encoder produces a contextual representation. We use the \texttt{[CLS]} representation followed by a linear classifier to obtain logits $z_i \in \mathbb{R}^{K}$. The same encoder and classifier are used for both the baseline and the candidate. The difference lies only in the training objective.

The baseline objective is label-smoothed cross-entropy,
\begin{equation}
\mathcal{L}_{\mathrm{LS}}=\mathrm{CE}_{\mathrm{LS}}(z_i,y_i;\epsilon),
\end{equation}
where $\epsilon=0.05$ in the reported experiments. Label smoothing is a strong and simple baseline because it discourages overconfident predictions without changing model architecture.

AutoTail first measures the imbalance of the training label distribution. Let $n_k$ be the number of training examples in class $k$ and $p_k=n_k/\sum_j n_j$. We compute normalized entropy,
\begin{equation}
H_{\mathrm{norm}}(p)=-\frac{\sum_k p_k\log(p_k+\epsilon_0)}{\log K},
\end{equation}
and define an imbalance index $I(p)=1-H_{\mathrm{norm}}(p)$. If $I(p)<\gamma$, prior correction is disabled. Otherwise, the method defines a tail set $\mathcal{T}$ as classes at or below the median non-zero class count. With smoothed priors
\begin{equation}
\pi_k=\frac{n_k+\alpha}{\sum_j n_j+\alpha K},
\end{equation}
the masked prior vector is
\begin{equation}
m_k=
\begin{cases}
\log \pi_k, & k\in\mathcal{T}\ \mathrm{and}\ I(p)\geq\gamma,\\
0, & \mathrm{otherwise}.
\end{cases}
\end{equation}
The adjusted logits are
\begin{equation}
\tilde{z}_{i,k}=z_{i,k}+\tau m_k.
\end{equation}
The reported configuration uses $\gamma=0.1$, $\alpha=1.0$, and $\tau=0.15$. The important design choice is gating: the prior is not applied indiscriminately to all classes and all tasks.

The second component is a weak Balanced Softmax auxiliary loss. Balanced Softmax incorporates class counts into the normalization term \citep{ren2020balanced}. In our setting it is used as an auxiliary signal rather than the main loss:
\begin{equation}
\mathcal{L}_{\mathrm{BSAux}}=\mathrm{CE}\left(z_i+\lambda_b\log(n+\alpha),y_i\right).
\end{equation}
The supervised objective becomes
\begin{equation}
\mathcal{L}_{\mathrm{sup}}=
\mathcal{L}_{\mathrm{LS}}(\tilde{z}_i,y_i) + w_b\mathcal{L}_{\mathrm{BSAux}},
\end{equation}
with $w_b=0.1$ and auxiliary temperature $0.25$.

The third component is FGM adversarial regularization. For embedding parameters $e$, FGM computes
\begin{equation}
r_{\mathrm{adv}}=\epsilon_{\mathrm{fgm}}\frac{\nabla_e \mathcal{L}_{\mathrm{sup}}}{\|\nabla_e \mathcal{L}_{\mathrm{sup}}\|_2}
\end{equation}
and optimizes an additional adversarial loss at $e+r_{\mathrm{adv}}$. The reported configuration uses $\epsilon_{\mathrm{fgm}}=0.3$ and adversarial loss weight $0.5$. FGM increases training time because it requires an additional perturbation step, but it does not add inference parameters.

The three components play different roles. AutoTail changes the class-prior geometry of the decision boundary, but only for classes that meet the tail criterion. The auxiliary Balanced Softmax term discourages the model from treating the observed label distribution as if it were the only meaningful deployment distribution. FGM then regularizes the representation space so that small embedding-level perturbations do not overturn the classification too easily. The method is therefore best understood as a training recipe for stabilizing imbalanced scholarly classifiers, not as a new encoder architecture.

\section{Experiment and Results}

\subsection{Experimental Setting}

We evaluate two base-size Chinese encoders: Chinese RoBERTa-WWM-ext and Chinese MacBERT-base. The first provides a strong Chinese whole-word-masking baseline; the second is a stronger sentence-level Chinese encoder and therefore a more demanding comparison. All neural models use \texttt{[CLS]} pooling, a linear classification head, batch size 16, learning rate $2\times10^{-5}$, weight decay 0.01, warmup ratio 0.1, and two fine-tuning epochs. The candidate and baseline use the same preprocessing, model size, seeds, and splits. No external private data, ensemble inference, larger-than-base encoder, or task-specific pretraining is used.

This design is conservative by intention. A method should not be credited for gains caused by a larger backbone, more data, or more inference-time computation. Conversely, the design also limits the maximum possible score. The paper therefore evaluates methodological usefulness, not leaderboard dominance.

The strongest comparison in the paper is MacBERT label smoothing versus MacBERT AutoTail-BSFGM. MacBERT is not the largest possible Chinese encoder, but it is a credible base-size backbone and a stronger baseline than a classical linear model or a weak BERT reproduction. RoBERTa-WWM is retained because it tests whether the method depends on one encoder family. If a method only improves a weak baseline, it is less interesting; if it improves both RoBERTa-WWM and MacBERT under the same task protocol, the evidence is stronger even if the absolute score remains below much larger systems.

\subsection{Main Abstract-to-Discipline Results}

Table~\ref{tab:main_results} reports three-seed results on the primary CSL abstract-to-discipline task. AutoTail-BSFGM improves all mean metrics for both encoders. With RoBERTa-WWM, validation accuracy rises from 67.28 to 68.69, while lockbox accuracy rises from 68.96 to 69.44. Macro-F1 and balanced accuracy also improve on both splits. With MacBERT-base, validation accuracy rises from 67.86 to 68.69, and lockbox accuracy rises from 69.41 to 69.90. These gains are smaller than the RoBERTa gains but remain positive under a stronger encoder.

\begin{table}[t]
\centering
\caption{Main three-seed results on CSL abstract-to-discipline classification. Scores are percentages. AT-BSFGM denotes AutoTail-BSFGM.}
\label{tab:main_results}
\scriptsize
\setlength{\tabcolsep}{2.2pt}
\begin{tabular}{@{}lllrrrrrrr@{}}
\toprule
Task & Encoder & Method & Val Acc & Val Macro & Val Bal. & Lock Acc & Lock Macro & Lock Bal. & Sec. \\
\midrule
Abstract & RoBERTa-WWM & Label smoothing & 67.28 & 64.59 & 64.13 & 68.96 & 65.10 & 64.64 & 2076 \\
Abstract & RoBERTa-WWM & AutoTail-BSFGM & 68.69 & 66.20 & 65.61 & 69.44 & 65.85 & 65.31 & 3290 \\
Abstract & MacBERT & Label smoothing & 67.86 & 65.09 & 64.59 & 69.41 & 65.71 & 65.02 & 1970 \\
Abstract & MacBERT & AutoTail-BSFGM & 68.69 & 65.98 & 65.43 & 69.90 & 66.30 & 65.55 & 3891 \\
Title & MacBERT & Label smoothing & 81.63 & 72.50 & 71.23 & 82.50 & 75.76 & 74.49 & 427 \\
Title & MacBERT & AutoTail-BSFGM & 82.33 & 73.83 & 73.87 & 82.42 & 76.02 & 75.71 & 680 \\
\bottomrule
\end{tabular}

\end{table}

\begin{figure}[t]
\centering
\includegraphics[width=\linewidth]{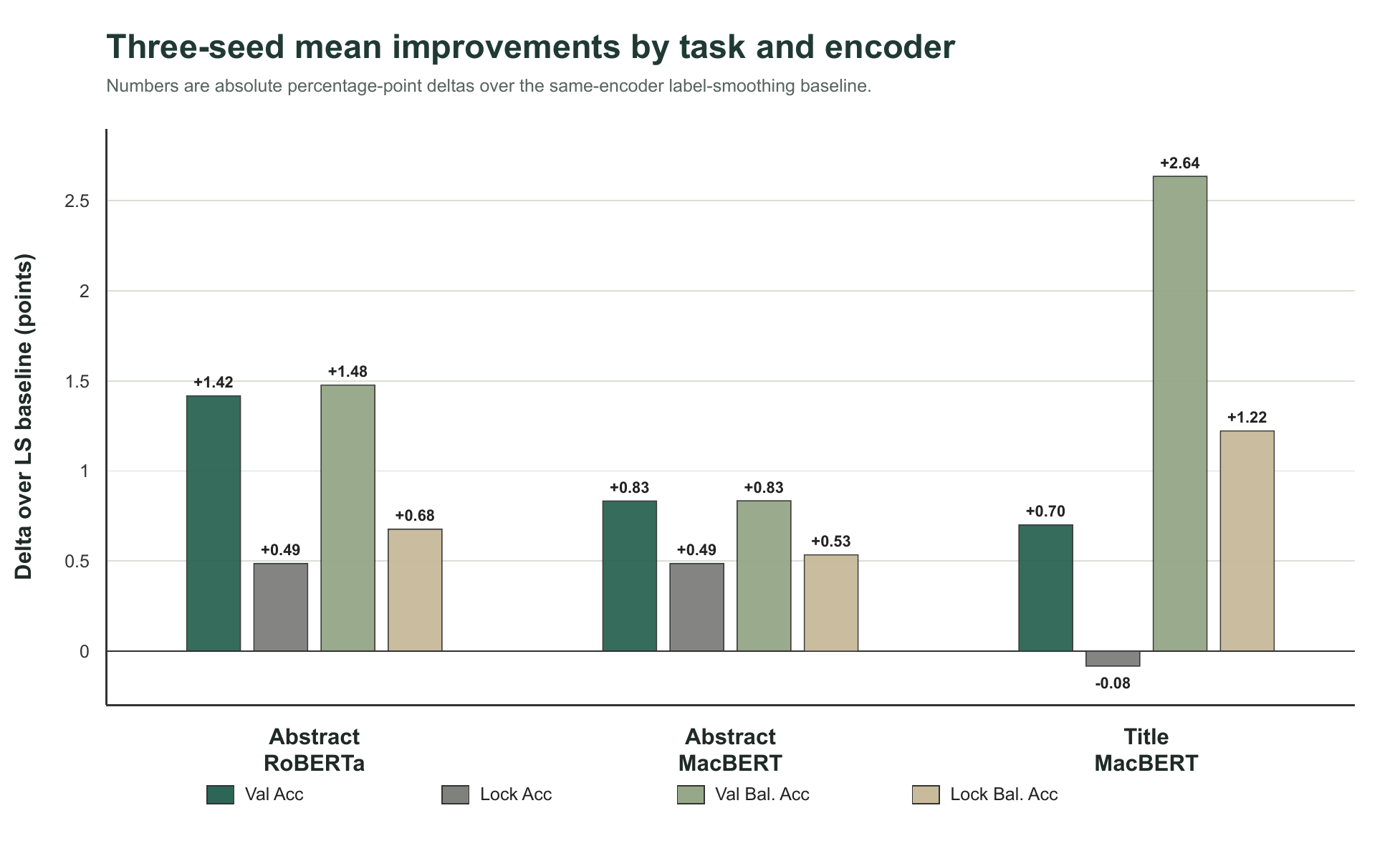}
\caption{Mean metric deltas over same-encoder label-smoothed baselines. Positive values indicate AutoTail-BSFGM improvement. The primary abstract task shows positive accuracy and balance-aware gains under both encoders; the title task shows a stronger balance-aware signal than raw lockbox accuracy.}
\label{fig:main_results}
\end{figure}

Figure~\ref{fig:main_results} presents the same evidence as deltas over the same-encoder label-smoothed baseline. The visualization makes two points clear. First, the primary abstract task has positive validation and lockbox deltas for both RoBERTa and MacBERT. Second, the title task has a more mixed profile: validation improves and balanced metrics improve, but lockbox accuracy is not consistently positive. This is precisely why the paper uses multiple metrics and a protected split.

\subsection{Cross-Task Title-to-Category Results}

The CSL title-to-category task changes the input from abstracts to titles and reduces the label set to 13 broad categories. This task is not designed to confirm the exact same behavior. Instead, it tests whether the balance-aware signal survives a different scholarly classification setting.

The results are mixed but informative. MacBERT AutoTail-BSFGM improves validation accuracy from 81.63 to 82.33, validation macro-F1 from 72.50 to 73.83, and validation balanced accuracy from 71.23 to 73.87. On lockbox, accuracy moves from 82.50 to 82.42, a small negative change of 0.08 points, while macro-F1 rises from 75.76 to 76.02 and balanced accuracy rises from 74.49 to 75.71. The most defensible interpretation is not that the method is universally better. Rather, it shifts the classifier toward more balanced category behavior, which can improve macro and balanced metrics even when overall accuracy is unchanged.

This result is valuable precisely because it is not perfectly positive. If the method had been evaluated only on validation accuracy, it would be easy to claim a clean transfer gain. The lockbox result blocks that interpretation. At the same time, the balanced-metric gains suggest that the method is not simply noise. It appears to redistribute some decision capacity across category frequencies. For scholarly classification, this may be desirable in workflows where minority fields matter, but it is not enough for a strict official benchmark claim.

For information science applications, this distinction matters. A research intelligence system may value minority-category recall and balanced treatment across fields, especially when the goal is to monitor emerging or underrepresented research areas. A leaderboard with accuracy as the only metric would judge the title lockbox result as neutral. A category-balance evaluation views it as potentially useful but not conclusive.

\subsection{Significance and Cost}

Table~\ref{tab:sig_cost} summarizes paired McNemar checks and training cost. On the MacBERT abstract validation split, the candidate correctly predicts 97 examples that the baseline misses, while the baseline correctly predicts 67 examples that the candidate misses; the exact McNemar $p$ value is 0.023. On the abstract lockbox split, the direction remains positive but is not statistically significant at the same threshold. On the title validation split, the direction is positive but weaker ($p=0.071$). On the title lockbox split, the evidence does not support an accuracy improvement.

\begin{table}[t]
\centering
\caption{Prediction-derived paired checks and training-cost ratios. McNemar counts are pooled over the three reported seeds for MacBERT comparisons.}
\label{tab:sig_cost}
\small
\setlength{\tabcolsep}{3.5pt}
\begin{tabular}{@{}llrrrrr@{}}
\toprule
Comparison & Split & Base only & Cand. only & Discord. & $p$ & Time x \\
\midrule
Abstract MacBERT & validation & 67 & 97 & 164 & 0.023 & 1.98 \\
Abstract MacBERT & lockbox & 64 & 78 & 142 & 0.275 & 1.98 \\
Title MacBERT & validation & 51 & 72 & 123 & 0.071 & 1.59 \\
Title MacBERT & lockbox & 47 & 45 & 92 & 0.917 & 1.59 \\
\bottomrule
\end{tabular}

\end{table}

Training cost is the main practical disadvantage. AutoTail-BSFGM increases training time by approximately 1.58$\times$ for the RoBERTa abstract task, 1.98$\times$ for the MacBERT abstract task, and 1.59$\times$ for the title task. This cost is expected because FGM requires an additional adversarial step. The cost does not affect inference. Therefore, the method is most appropriate when training-time overhead is acceptable and when category balance matters enough to justify that overhead.

The cost should be interpreted in relation to the deployment scenario. For a digital library that retrains a classifier periodically and then serves predictions for a large corpus, inference cost is usually more important than one-time training cost. In that setting, AutoTail-BSFGM is attractive because deployment remains unchanged. For an interactive system that must retrain frequently on small user-defined taxonomies, the additional training time may be less acceptable. This cost-benefit distinction is part of why the paper avoids describing the method as universally preferable.

\section{Case Study: Category-Level Behavior}

Aggregate accuracy can obscure where a method improves. To examine category-level behavior, we group labels by training frequency into tail, mid, and head bands and compute prediction accuracy within each band. The grouping is not meant to replace per-class analysis; it is a compact diagnostic for whether gains concentrate only in large categories or also affect underrepresented labels.

\begin{figure}[t]
\centering
\includegraphics[width=\linewidth]{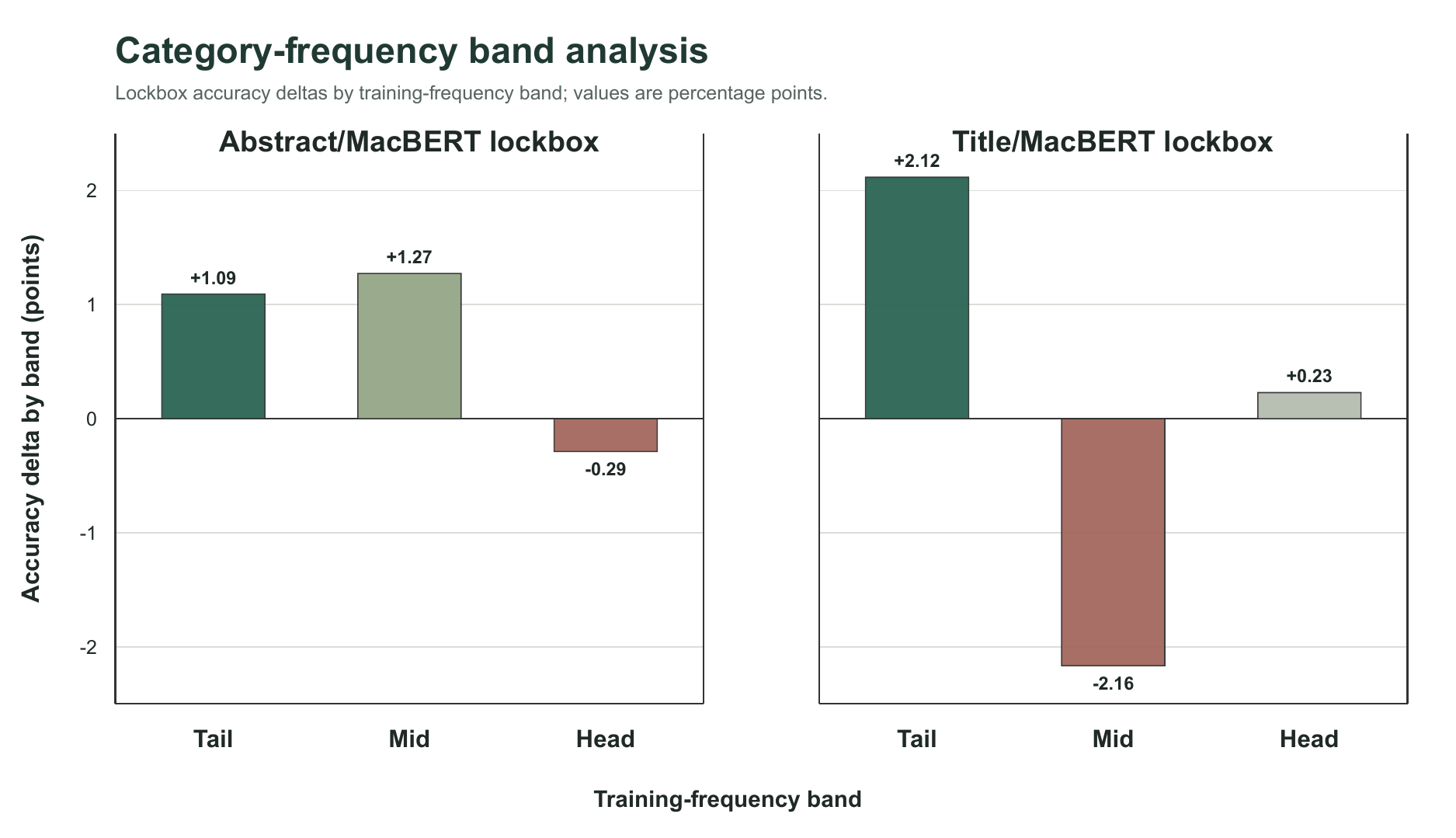}
\caption{Category-band accuracy deltas for MacBERT comparisons, grouped by training-label frequency. The abstract task shows positive tail and mid-band movement on both validation and lockbox splits. The title task has more mixed head-band behavior, explaining why balanced metrics improve while lockbox accuracy remains neutral.}
\label{fig:category_delta}
\end{figure}

Figure~\ref{fig:category_delta} shows category-band deltas for the MacBERT comparisons. On the abstract validation split, AutoTail-BSFGM improves tail, mid, and head bands, with the largest gain in the mid band. On the abstract lockbox split, the tail band improves by about 1.09 points, the mid band improves by about 0.85 points, and the head band is approximately neutral. This is consistent with the method's design: it does not simply boost all tail logits aggressively, but it changes the training dynamics enough to improve non-head performance without a major head-class penalty.

\begin{table}[t]
\centering
\caption{Category-band analysis derived from prediction files. Examples are counted after pooling three seeds, so each validation or lockbox example contributes once per seed.}
\label{tab:category_delta}
\small
\setlength{\tabcolsep}{3.5pt}
\begin{tabular}{@{}lllrrrr@{}}
\toprule
Task & Split & Band & N x seed & Base Acc & Cand. Acc & $\Delta$ \\
\midrule
Abstract/MacBERT & validation & Tail & 684 & 53.36 & 54.09 & 0.73 \\
Abstract/MacBERT & validation & Mid & 1182 & 71.15 & 72.42 & 1.27 \\
Abstract/MacBERT & validation & Head & 1734 & 71.34 & 71.91 & 0.58 \\
Abstract/MacBERT & lockbox & Tail & 549 & 53.37 & 54.46 & 1.09 \\
Abstract/MacBERT & lockbox & Mid & 942 & 68.05 & 69.32 & 1.27 \\
Abstract/MacBERT & lockbox & Head & 1389 & 76.67 & 76.39 & -0.29 \\
Title/MacBERT & validation & Tail & 282 & 63.83 & 69.86 & 6.03 \\
Title/MacBERT & validation & Mid & 507 & 73.57 & 71.99 & -1.58 \\
Title/MacBERT & validation & Head & 2211 & 85.75 & 86.30 & 0.54 \\
Title/MacBERT & lockbox & Tail & 189 & 73.54 & 75.66 & 2.12 \\
Title/MacBERT & lockbox & Mid & 462 & 75.11 & 72.94 & -2.16 \\
Title/MacBERT & lockbox & Head & 1749 & 85.42 & 85.65 & 0.23 \\
\bottomrule
\end{tabular}

\end{table}

The title task explains the main limitation. Validation gains are visible in tail and mid categories, but lockbox accuracy does not improve overall. This pattern suggests that the method can change the class-balance profile without guaranteeing a universal accuracy improvement. For applied scholarly systems, such behavior may still be useful if the objective is balanced coverage. For a strict benchmark claim, it is insufficient unless the headline metric also improves on the protected split.

The category-level view also offers a diagnostic for future method development. If the method improves tail categories only by harming head categories, it may be a simple reweighting artifact. If it improves head categories only, it is not solving the balance problem. The abstract task is promising because gains are visible in tail and mid bands while head performance remains roughly stable. The title task is less stable because mid-band behavior declines in the lockbox analysis even though tail and head bands improve. This suggests that the next iteration should focus on semantic confusion among adjacent middle-frequency categories rather than increasing the global tail prior.

The case study also clarifies why the method should not be described as a general-purpose state-of-the-art algorithm. A stronger claim would require larger task coverage, more seeds, additional baselines such as direct Balanced Softmax, logit adjustment, FGM-only, and stronger encoders, and ideally an external test submission. The present evidence is better framed as a reproducible balance-aware fine-tuning result with promising but bounded generalization.

For a journal article, the most important future evidence would be error-typed examples. The present package records prediction-derived band results, but it does not yet include a qualitative taxonomy of errors such as method-domain confusion, application-domain ambiguity, cross-disciplinary overlap, or insufficient title evidence. Such an analysis would connect the machine-learning result more directly to information science. It would also help decide whether a balance-aware classifier actually improves the work of human analysts or simply moves numerical metrics.

\section{Conclusions and Implications}

\subsection{Practical Implications}

For digital libraries and research intelligence tools, the main practical message is that small training-objective changes can alter category-level behavior without changing inference infrastructure. AutoTail-BSFGM keeps the same base encoder and linear classifier at deployment time. This makes it easy to integrate into systems that already use Chinese pretrained encoders for classification. The method is most relevant when underrepresented categories matter, such as monitoring emerging disciplines, constructing discipline-level dashboards, or reducing head-category dominance in literature organization.

The validation-lockbox design is also practically important. If a system is used for research assessment or knowledge organization, a single development-set gain is not enough. The title task shows why: validation accuracy improves, but lockbox accuracy is neutral. A deployment-oriented workflow should therefore maintain a protected split, inspect macro and balanced metrics, and audit class-level errors before claiming improvement.

\subsection{Technical Implications}

Technically, the results suggest that prior-aware and robustness-aware mechanisms can be complementary in scholarly classification. AutoTail alone encodes a class-prior assumption; Balanced Softmax supplies a weak frequency-aware auxiliary signal; FGM regularizes local embedding sensitivity. The combined method is not a large architectural innovation, but its value lies in making the correction conditional and auditable. Gating the prior adjustment reduces the risk of applying a long-tail correction where the distribution does not warrant it.

The evidence also cautions against interpreting balance-aware methods through accuracy alone. On the title task, balanced accuracy improves more than raw accuracy. This indicates that future evaluation of scholarly classifiers should report multiple metrics and analyze category bands. In disciplinary classification, minority labels are not merely statistical noise; they may correspond to fields that are important for discovery and policy analysis.

\subsection{Limitations and Future Directions}

This study has several limitations. First, the evidence is limited to two CSL-based tasks and three seeds. Although the abstract task shows consistent gains under two encoders, broader validation is required before making a general claim. Second, the method increases training time by roughly 1.6--2.0$\times$, primarily because of FGM. This cost may be acceptable for offline scholarly indexing but less attractive for frequent retraining. Third, the current study does not include official leaderboard submission or external institutional data. The lockbox split reduces overfitting risk, but it is not a substitute for external validation.

Fourth, the component analysis is not exhaustive. The reported configuration emerged from constrained experimentation, but a full paper should include systematic ablations for AutoTail-only, Balanced-Softmax-only, FGM-only, different prior temperatures, and alternative robust fine-tuning methods such as FreeLB or SMART. Fifth, the category-band analysis groups labels by frequency rather than semantic relations. Recent prompt and long-tail text-classification methods suggest that label verbalizers, label-document retrieval, hierarchical taxonomies, and LLM-assisted tail augmentation can supply information that class counts alone cannot capture \citep{yu2024sciprompt,wang2023pesco,lu2026dealt}. Future work should examine confusion between related disciplines, use hierarchical taxonomies where available, and connect classification errors to downstream retrieval or research-intelligence decisions.

The next research step is therefore not to claim a universal breakthrough, but to strengthen the evidence boundary. A journal-ready extension should add more scholarly classification tasks, stronger ablations, official test submissions where possible, and qualitative error analysis with representative examples. It should also test whether AutoTail-BSFGM can be combined with label semantics or LLM-assisted tail augmentation without losing the auditability of the current single-encoder protocol. If the balance-aware gains persist under those checks, AutoTail-BSFGM can be positioned as a practical method for robust scholarly text classification rather than as a narrow score-tuning trick.

\section*{Data Availability Statement}

The experiments use CSL-derived public data and locally generated train, validation, lockbox, and test splits. The accompanying \href{https://github.com/thu-nmrc/autotail-bsfgm-scholarly-classification}{GitHub repository} provides the paper source, generated CSV tables, editable figure sources, run identifiers, experiment configurations, a reproducibility guide, and an artifact manifest indicating the local prediction and run-registry files used to derive the reported values. Model checkpoints and large prediction-level artifacts are not included in the source repository, but they can be regenerated from the recorded configurations.

\section*{Conflict of Interest Statement}

The authors report no conflict of interest for this arXiv version.

\bibliographystyle{plainnat}
\bibliography{references}

\end{document}